\documentclass[10pt,twocolumn,letterpaper]{IEEEtran}
\usepackage{framed,multirow}
\usepackage{amssymb}
\usepackage{latexsym}
\usepackage{subfigure}
\usepackage{amsmath}
\usepackage{chngcntr}
\usepackage{cite}
\newcommand{\tabincell}[2]{\begin{tabular}{@{}#1@{}}#2\end{tabular}}
\usepackage{url}
\usepackage{xcolor}
\definecolor{newcolor}{rgb}{.8,.349,.1}
\usepackage[colorinlistoftodos]{todonotes}  
\def\BibTeX{{\rm B\kern-.05em{\sc i\kern-.025em b}\kern-.08emT\kern-.1667em\lower.7ex\hbox{E}\kern-.125emX}}
\usepackage[colorlinks,linkcolor=blue,anchorcolor=blue,citecolor=blue]{hyperref} 
\usepackage{authblk}
\usepackage{geometry}
\usepackage{url}
\usepackage{amsmath}
\usepackage{multirow}

\newcommand{\etal}{\emph{et al. }}       

\newcommand{\eg}{\emph{e.g. }}

\begin{document}
\title{DIODE: Dilatable Incremental Object Detection}
\author{Can Peng}
\author{Kun Zhao}
\author{Sam Maksoud}
\author{Tianren Wang}
\author{Brian C. Lovell}
\affil{School of ITEE, The University of Queensland, Brisbane, QLD, Australia}
\date{}
\maketitle
		
\begin{abstract}
To accommodate rapid changes in the real world, the cognition system of humans is capable of continually learning concepts.
On the contrary, conventional deep learning models lack this capability of preserving previously learned knowledge.
When a neural network is fine-tuned to learn new tasks, its performance on previously trained tasks will significantly deteriorate.
Many recent works on incremental object detection tackle this problem by introducing advanced regularization.
Although these methods have shown promising results, the benefits are often short-lived after the first incremental step.
Under multi-step incremental learning, the trade-off between old knowledge preserving and new task learning becomes progressively more severe.
Thus, the performance of regularization-based incremental object detectors gradually decays for subsequent learning steps.
In this paper, we aim to alleviate this performance decay on multi-step incremental detection tasks by proposing a dilatable incremental object detector (DIODE).
For the task-shared parameters, our method adaptively penalizes the changes of important weights for previous tasks.
At the same time, the structure of the model is dilated or expanded by a limited number of task-specific parameters to promote new task learning.
Extensive experiments on PASCAL VOC and COCO datasets demonstrate substantial improvements over the state-of-the-art methods.
Notably, compared with the state-of-the-art methods, our method achieves up to 6.0\% performance improvement by increasing the number of parameters by just 1.2\% for each newly learned task.
\end{abstract}

\section{Introduction}
\label{sec: Introduction}
	
In recent years, benefiting from the development of deep learning, the computer vision community has witnessed incredible performance improvements on many tasks, such as image classification and object detection.
In addition to the theoretical development of deep neural networks and the advent of GPUs, the vast amount of accessible public image data is another key reason for these performance breakthroughs.
Normally, to train an accurate neural network, a large amount of correctly labeled data for predefined categories is required to supervise the learning procedure.
However, for real-life scenarios, these training data usually arrive sequentially instead of being provided as a block before training.
In many practical situations, where only new task data are provided, the ideal neural network model would continually learn multiple tasks without forgetting previously learned tasks.
In reality, if the neural network is directly fine-tuned with only new task data, it rapidly forgets what it has learned before --- a problem called catastrophic forgetting \cite{goodfellow2013empirical, mccloskey1989catastrophic}.
Incremental learning research focuses on addressing this problem.

\begin{figure}[tbp]
	\centering
	\includegraphics[width=8cm, keepaspectratio]{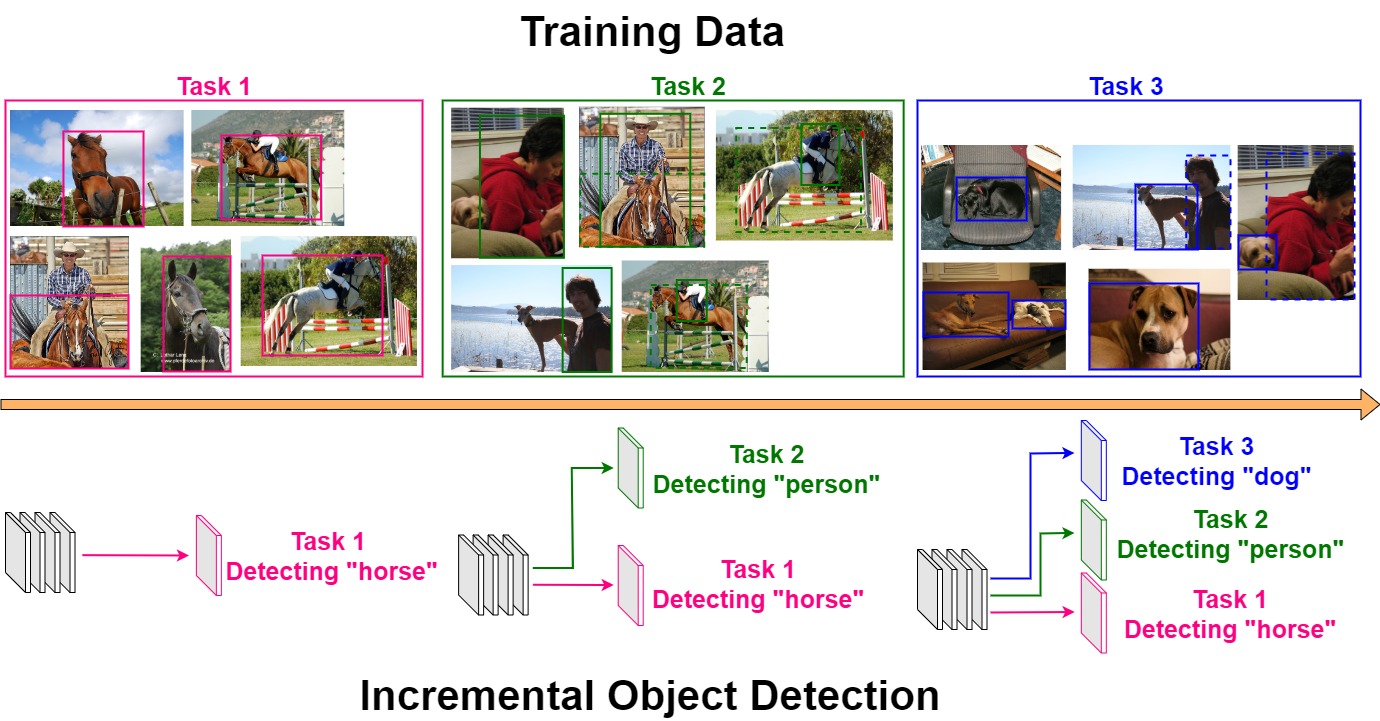}
	\caption{An example of an incremental object detection task.
		The detection model is required to learn to detect `horse', then `person', and then `dog' objects.
		Solid boxes on the training data images are the provided annotations for the new class objects.
		Dotted boxes on the images are the missed annotations for the old class objects.
		Incremental object detection is a challenging task since the model not only suffers from catastrophic forgetting but additionally suffers from missing annotations for old task objects in new task data.}
	\label{fig: incremental_object_dectection}
\end{figure}
	
In this paper, we tackle incremental object detection which is a very important but relatively less explored task than incremental classification.
Figure \ref{fig: incremental_object_dectection} shows an example of incremental object detection.
Compared with incremental classification, incremental object detection is more challenging since the model not only suffers from catastrophic forgetting but also from missing annotations for old task objects in the new task data.
Under multi-step incremental detection scenarios, the difficulties of memorizing old classes while learning new classes will exponentially accumulate due to both catastrophic forgetting and missing annotations.
Compared to one-step incremental learning, multi-step incremental learning commonly occurs in real life, since there might be new types of objects arising at anytime and anywhere.
Currently, most of the research on incremental object detection use additional regularization such as Knowledge Distillation \cite{hinton2015distilling} (KD) or Elastic Weight Consolidation (EWC) \cite{kirkpatrick2017overcoming} to alleviate the catastrophic forgetting effect.
Although these methods have shown promising results on benchmark datasets, performance improvement is short-lived --- the methods show good performance for the first incremental step but performance rapidly decays during the following steps.
This is because, with a fixed network structure, the extra regularization methods suffer from a conflict between preserving old knowledge and accommodating new knowledge.

\begin{figure}[tbp]
	\centering
	\includegraphics[width=8cm, keepaspectratio]{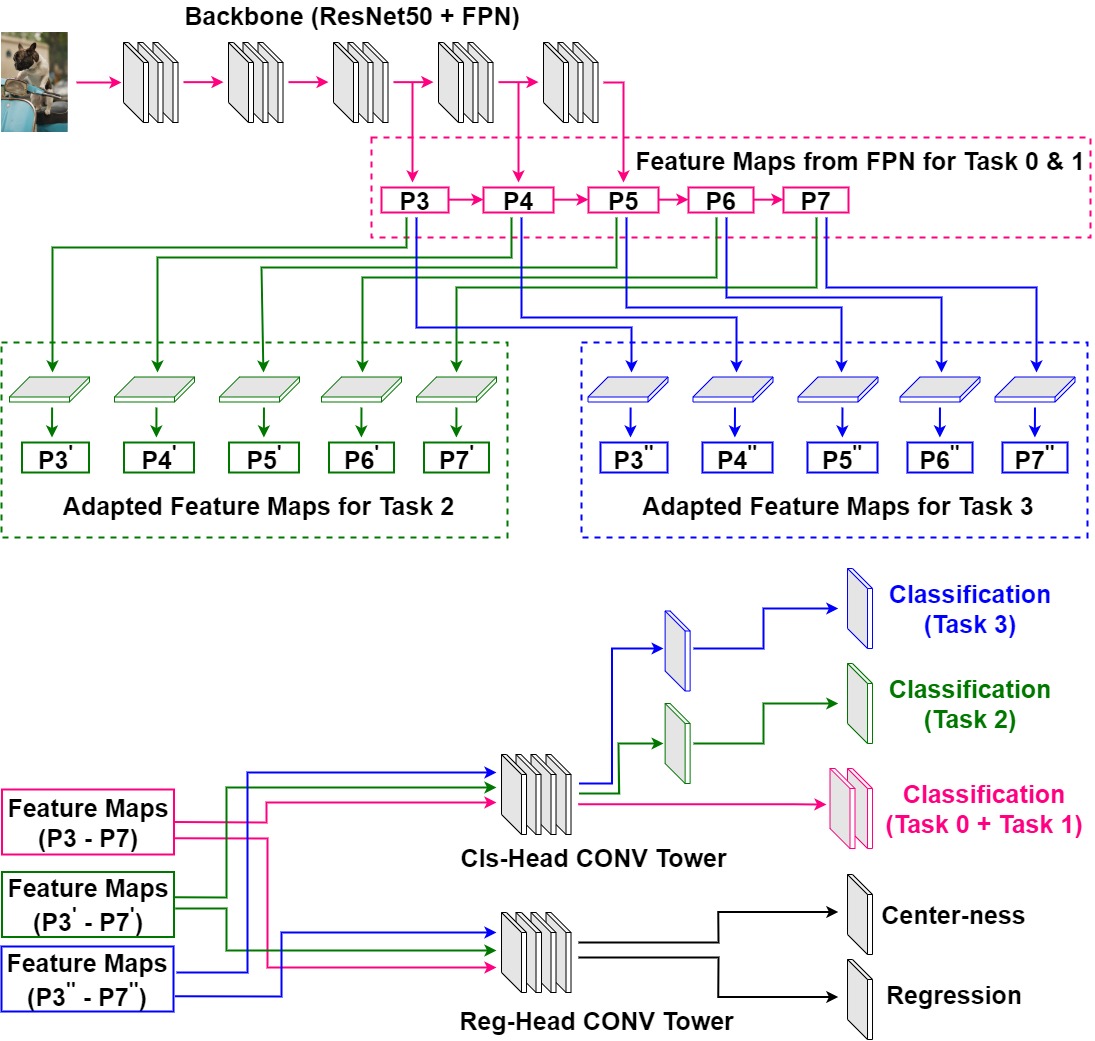}
	\caption{The framework of the proposed DIODE method.
		Task 0 is the initial task that is normally trained.
		Tasks 1, 2, and 3 are the incrementally learned tasks.
		For each step of incremental learning, constrained EWC is applied to preserve old knowledge.
		Also, to promote new task learning as well as make full use of existing parameters, from the second step of incremental learning, task-specific dilatable modules are assigned for each task.}
	\label{fig: DIODE_backbone}
\end{figure}
	
To solve the challenging multi-step incremental detection task, here we propose a dilatable incremental object detector (DIODE) based on constrained EWC.
Figure \ref{fig: DIODE_backbone} illustrates the framework of our proposed DIODE method on the anchor-free fully convolutional one-stage object detector (FCOS) \cite{tian2019fcos}.
As neural networks are usually designed with some superfluous parameters, we do not expand the model structure until the second step of incremental learning.
For the first step of incremental learning (task 1), during the new task learning, the constrained EWC method is applied to the backbone of the detector to preserve old task knowledge.
For the following incremental learning steps (\eg task 2, task 3), the constrained EWC is still applied to the backbone of the detector while the feature maps are adapted by adding task-specific filters to promote new task learning.
\textbf{The contributions} of this paper are as follows:
\begin{itemize}
	\item We analyze the EWC-based incremental classification method and adapt the approach to our incremental object detection task.
	\item We propose a dilatable incremental object detector (DIODE), which leverages the strengths of EWC as well as dynamic architecture methodologies to simultaneously improve incremental detection performance.
	\item To the best of our knowledge, we are the first to explore incremental object detection with a dilatable network structure.
	\item The experiments on benchmark datasets (VOC and COCO) demonstrate that the proposed method performs significantly better than the current state-of-the-art incremental methods on multi-step detection tasks.
\end{itemize}
The paper is organized as follows: Section \ref{sec: Related Work} introduces the related work on incremental learning and Section \ref{sec: Methodology} describes the proposed method in detail.
Experimental results are shown in Section \ref{sec: Experiments} and Section \ref{sec: Conclusion} concludes the paper.
	
\section{Related Work}
\label{sec: Related Work}
In this section, we introduce the state-of-the-art methods for incremental classification followed by methods for incremental object detection.
	
\subsection{Incremental Classification}
Methods for incremental classification usually adopt three types of strategies: regularization, architectural, and rehearsal.
	
\noindent\textbf{Regularization Strategies} alleviate the catastrophic forgetting problem by imposing constraints on the update of the network parameters.
This is generally achieved by additional regularization terms on the loss function and can be further divided into prior-focused and data-focused methods.
Prior-focused approaches penalize the changes of important weights for old tasks in order to preserve the old task knowledge.
Elastic Weight Consolidation (EWC) \cite{kirkpatrick2017overcoming} introduced by Kirkpatrick \etal in 2016 was the first to establish the prior-focused regularization approach.
Inspired by the EWC method, some works followed this line of research and the differences lie in the methods used to decide the importance of each weight \cite{aljundi2018memory, zenke2017continual}.
In addition, methods based on node-wise importance have also been explored.
Jung \etal proposed two group sparsity-based penalties on the loss function and selectively employed the two penalties during new task learning according to the importance of the nodes towards the old tasks \cite{jung2020continual}.
In contrast to this, data-focused approaches penalize changes in the input-output function of the neural network for the old tasks by using new task data to approximate the performance of the previous tasks.
Knowledge Distillation (KD) \cite{hinton2015distilling} is the basic building block in data-focused methods.
Li \etal first applied KD for incremental learning and built an incremental classifier called Learning without Forgetting (LwF) \cite{li2017learning}.
Research in this type of method aims to acquire the old task information through various distillation approaches \cite{zhao2020maintaining, zhou2019m2kd}.
	
\noindent\textbf{Architectural Strategies} solve the catastrophic forgetting problem by designing a unique sub-network for each incremental task.
Fernando \etal proposed PathNet that uses agents embedded in the neural network to discover which part of the network is suitable to be re-used for new tasks \cite{fernando2017pathnet}.
Inspired by network pruning techniques, Mallya \etal proposed PackNet \cite{mallya2018packnet} which iteratively assigns parameter subsets to consecutive tasks by constituting binary masks.
However, with a fixed network structure, the spare neurons will eventually be exhausted and the network will become rigid towards new task learning.
To solve this, researchers explored dynamic architectures which expand the model structure during learning.
Rusu \etal proposed a method called Progressive Neural Network (PNN) which blocks any change towards the parameters trained on the old tasks and continually expands the architecture by allocating new sub-networks with fixed capacity for the new tasks \cite{rusu2016progressive}.
Rosenfeld \etal proposed Deep Adaptation Network (DAN) that constrains newly learned filters to be linear combinations of existing ones \cite{rosenfeld2018incremental}.
Abati \etal equipped each convolutional layer within the network with task-specific gating modules to select which filters to be applied on the given input \cite{abati2020conditional}.
Singh \etal mentioned that the activation maps of the neural network trained on the old task can be calibrated using limited calibration parameters and then become relevant to the new task \cite{singh2020calibrating}.
	
\noindent\textbf{Rehearsal Strategies} store some old task samples, and during new task learning, these old task samples are replayed alongside the new task data to avoid catastrophic forgetting.
Rebuffi \etal introduced a KD-based incremental classifier called iCaRL \cite{rebuffi2017icarl}.
Different from LwF, iCaRL stores some representative old class exemplars which are selected through herding \cite{rebuffi2017icarl}.
The old class exemplars and new data are combined to train the new model.
However, as only limited exemplars are stored, imbalanced cardinality between the old and new classes causes a prediction bias towards the new classes.
Many efforts have been aimed at solving this bias problem \cite{castro2018end, hou2019learning, wu2019large}.
	
\subsection{Incremental Object Detection}
Currently, almost all of the incremental object detection methods are inspired by regularization strategies for incremental classification.
Using Fast RCNN \cite{Girshick2015} as the backbone network, Shmelkov \etal proposed the first KD-based incremental detector ILOD \cite{shmelkov2017incremental}.
Following Shmelkov \etal, many researchers have designed various KD-based incremental detectors on more advanced object detection architectures.
Hao \etal \cite{Hao2019AnEA}, Chen \etal \cite{8851980} and Peng \etal \cite{peng2020faster} designed KD-based incremental detectors based on Faster R-CNN \cite{ren2015faster}.
Li \etal \cite{li2019rilod} and Zhang \etal \cite{zhang2020class} proposed KD-based incremental detectors based on RetinaNet \cite{lin2017focal}.
Perez \etal designed a KD-based few-shot incremental detector \cite{perez2020incremental} on CenterNet \cite{zhou2019objects}.
Liu \etal proposed IncDet \cite{liu2020incdet} based on EWC and used the Huber regularization to clip the gradient of each parameter according to its contribution to the old classes.
	
Although regularization strategies are popular for incremental learning and have shown promising performance on detection tasks, there are some serious limitations.
With a limited amount of neural resources, the catastrophic forgetting of old knowledge and the inability to learn new knowledge leads to a stability-plasticity trade-off problem.
The old task knowledge is preserved at the cost of impeding new task learning.
For long-range multi-step incremental learning, with a fixed model capability, regularization approaches often suffer a significant performance decrease in later tasks.
On the other hand, for dynamic architecture strategies, they normally fix the parameters learned for the old tasks which may make the final model suffer from a cumbersome network structure.
Considering the complementary advantages and disadvantages of regularization and architectural approaches, we try to unify these techniques in one framework.
In this paper, we propose to design a hybrid incremental detector that can maximally utilize each neuron of the existing model without destroying the important parameters for the old tasks.
Our hybrid method will learn the incremental difference for the new tasks through the addition of a modest number of newly added parameters.
	
\section{Methodology}
\label{sec: Methodology}
	
In this section, using FCOS \cite{tian2019fcos} as the backbone detector, we propose a dilatable network architecture with constrained regularization for incremental object detection.
First, we analyze the network structure of the FCOS detector.
To solve the missing annotation problem on the new task data, we use a simple but effective pseudo annotation strategy.
Then we apply a constrained regularization loss on task-shared parameters of the model to preserve the old task knowledge.
After that, we expand the network structure with modest new task-specific parameters to overcome the downsides of regularization and further foster new task learning.
	
\subsection{Backbone Detector - FCOS}
\label{sec: Backbone Detector - FCOS}
Using FCOS \cite{tian2019fcos} as the backbone network, we outline our proposed method.
Figure \ref{fig: DIODE_backbone} shows the framework of our method.
FCOS predicts each object as a point inside the object and regresses the distance from that point to the four sides of the bounding box.
ResNet \cite{he2016deep} and Feature Pyramid Network (FPN) \cite{lin2017focal} are used in FCOS as the backbone structure.
After generating multi-level features from FPN, the features are sent to two sets of convolutional lamination (`Cls-Head CONV Tower' and `Reg-Head CONV Tower') to generate the final features.
Then, the features from `Cls-Head CONV Tower' are sent to the `Classification' head to generate classification results.
Simultaneously, the outputs from `Reg-Head CONV Tower' are sent to `Center-ness' and `Regression' heads to generate center-ness and bounding box regression results, respectively.
	
\subsection{Pseudo Annotation}
\label{sec: Pseudo Annotation}
During training, the missing annotations for the old objects will misguide the detector to regard the old class objects as background and accelerate forgetting of old knowledge.
One very simple but effective way to mitigate the missing annotation problem is to use the detection results of the old model as pseudo annotations for old class objects on the new task data.
The pseudo annotations also alleviate catastrophic forgetting since they provide some old task information.
According to the experimental results in Section \ref{sec: Experiments}, even only using directly fine-tuning, with the help of pseudo annotations, model performance is greatly improved.
Thus, in our method, similar to \cite{liu2020incdet}, we use two sets of labels to train the incremental detector:
1) new class annotations provided from the new training data, and
2) old class annotations generated from the old model.
	
\subsection{Constrained EWC Module}
\label{sec: Constrained EWC Module}
Neuroscience suggests that the mammalian brain can avoid catastrophic forgetting by protecting previously acquired knowledge in neocortical circuits.
When mammals learn a new skill, a proportion of excitatory synapses are strengthened.
Then a synaptic consolidation mechanism will enable incremental learning by reducing the plasticity of the synapses that are vital to previously learned tasks\cite{hayashi2015labelling, yang2009stably}.
Inspired by this synaptic consolidation mechanism, Kirkpatrick \etal introduced the EWC algorithm \cite{kirkpatrick2017overcoming} to slow down learning on certain weights based on how important they are to previously seen tasks.
This means that these important parameters will stay close to their old values.
If a model has been properly trained on $task\textrm{ }1$ to $task\textrm{ }N$ and needs to continually learn the $task\textrm{ }N+1$ with only training data for $task\textrm{ }N+1$, the loss function used in EWC for incremental learning $task\textrm{ }N+1$ is:
\begin{small}
	\begin{equation}
	\mathcal L(\theta) = \mathcal L_{N+1}(\theta) \textrm{ }+\frac{\lambda}{2}\textrm{ }\sum_i(\sum_{n=1}^N F_{n,i})(\theta_i \textrm{ }-\textrm{ } \theta^\star_{N, i})^2
	\label{eq: EWC}
	\end{equation}
\end{small}
	
\noindent where $\mathcal L_{N+1}(\theta)$ is the normal training loss for learning $task\textrm{ }N+1$.
$\lambda$ is the hyper-parameter which balances the new task training loss and EWC regularization loss terms.
$F$ is the importance matrix which decides the importance of each parameter and EWC uses Fisher Information Matrix \cite{amari1998natural} to approximate parameter importance.
Following the online EWC method \cite{schwarz2018progress}, under multi-step incremental scenarios, the importance matrices for all previous tasks ($task\textrm{ }1$ to $task\textrm{ }N$) are accumulated to generate the final importance matrix for the training of  $task\textrm{ }N+1$.
In addition, $\theta$ is the set of model parameters, $\theta^\star$ is the final optimal parameter set for the last task ($task\textrm{ }N$), where $i$ indexes each parameter.
	
When directly adapting the EWC method to incremental object detection, we find that a gradient explosion usually happens and causes the unstable training.
A similar phenomenon has also been observed by Liu \etal and they claimed that due to the distributional shifts of images across different classes, certain parameters have extremely large gradients which leads to gradient explosion \cite{liu2020incdet}.
On their IncDet method \cite{liu2020incdet}, to avoid the explosion, the gradients of certain parameters which have extremely large values on the EWC regularization loss term are clipped via a threshold.
During EWC training, we analyze the importance matrix and find that:
1) The importance for different parameters has a huge difference, sometimes the gap can be ten up to powers of tens.
2) The parameters which have extremely large importance are usually parameters from the bounding box regression head or classification head.
	
According to our analysis on the EWC importance matrix, we conjecture that only applying EWC regularization to the feature extractor (ResNet, FPN, and `Cls-Head CONV Tower' for FCOS) can make it more effective for incremental object detection tasks.
The reasons are four-fold:
1) To keep the old task information, the class-wise parameters on the classification head can be directly fixed instead of EWC regularized during new task learning.
2) For anchor-free fully convolutional object detectors, such as FCOS, the regression outputs are partly trained only at positive pixels where objects exist.
The negative pixels which do not contain objects are not included in regression output training and will automatically follow the old model prediction since they are initialized by the model parameters which are not retrained.
Thus, omitting EWC regularization on the parameters of the bounding box regression head will merely affect regression on old class objects.
3) Since the parameters from classification and regression heads have extremely large importance, omitting them in EWC regularization can alleviate the gradient explosion problem.
4) Cropping the importance matrix to remove extremely large importance from the detection heads can make the EWC method focus more on the class-agnostic feature extractor and better preserve features important for the old tasks.
	
\subsection{Dilatable Module}
\label{sec: Dilatable Module}

\begin{figure}[tbp]
	\centering
	\includegraphics[width=8cm, keepaspectratio]{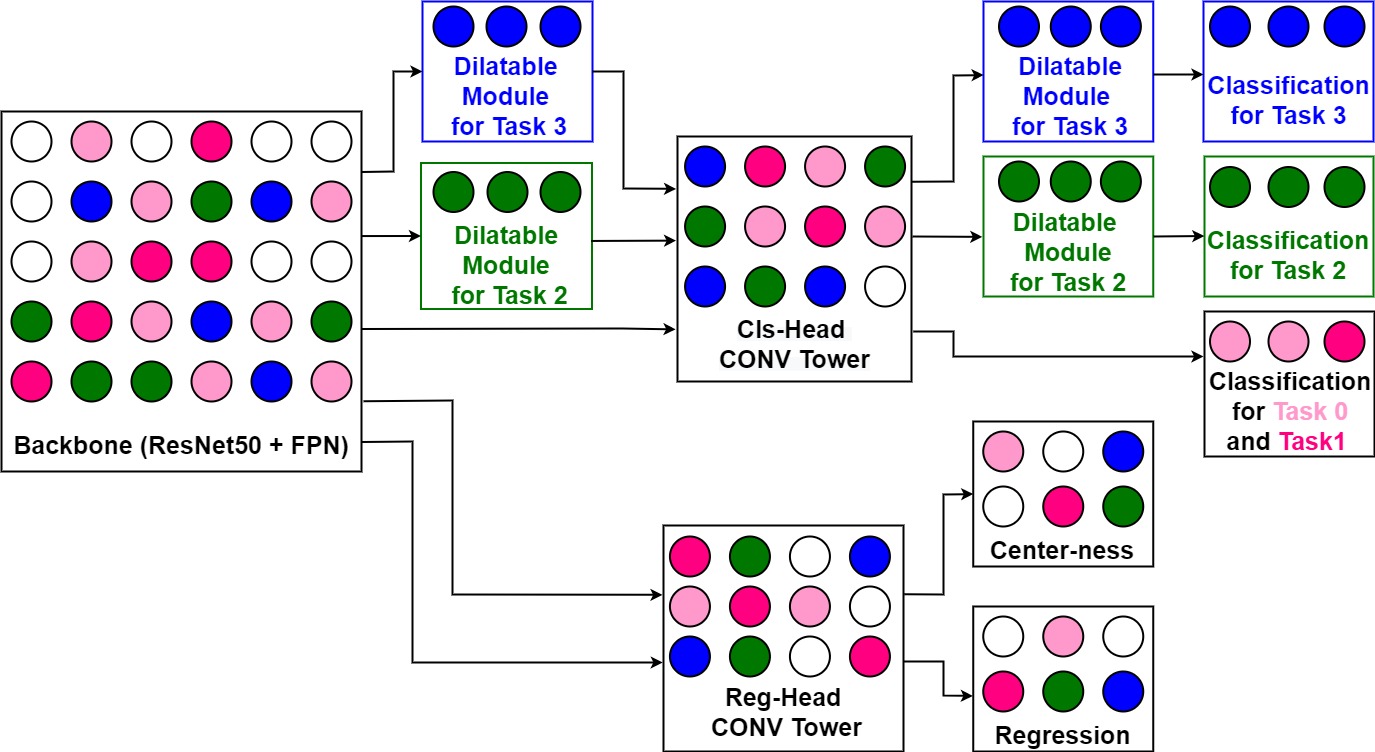}
	\caption{An example of the parameter distribution of the DIODE method.
		Each circle represents a neuron in the model.
		Task 0 is the initial task that is normally trained.
		Task 1, 2, and 3 are the incrementally learned tasks.
		Neurons within the backbone network are shared between different tasks and constrained EWC is used to maintain old task information.
		From the second step of incremental learning, dilatable modules are assigned for each new task to promote the learning of new knowledge.}
	\label{fig: DIODE_parameter}
\end{figure}

With the help of the EWC regularization, the updating of parameters that are important to old tasks is constrained, so the old task knowledge is preserved.
However, under realistic conditions, new type objects can emerge at any time and new task learning may be required indefinitely.
With a fixed neural network architecture, spare neurons will gradually be exhausted, and the network will eventually become rigid with respect to new task learning.
One solution to avoid rigidity is to expand the model structure.
Thus, while applying EWC regularization on the feature extractor, extra task-specific parameters are assigned to each new task be learned.
On the one hand, EWC regularization helps to maintain vital features for old task performance.
On the other hand, spare neurons (with low EWC importance) and newly added neurons help to learn features that are critical to new task performance.
Figure \ref{fig: DIODE_parameter} demonstrates the condition of the model parameters for our method.

We use one convolutional layer with kernel size one as the dilatable module.
Although the model structure is expanded whenever a new task comes in, the model structure will not become cumbersome as only a few new parameters are required.
Following the network architecture of FCOS, a new convolutional layer with kernel size three is also added for new class classification.
	
Assume $\mathcal X$ is the input image, $\mathcal F$ is the feature sent to `Cls-Head CONV Tower' layers, and $\mathcal Y$ is the classification output.
$CONV_{FPN}$, $CONV_{CH}$ and $CONV_{Cls}$ are the convolution layers for FPN, `Cls-Head CONV Tower' and `Classification', respectively.
Note that for each input image, the FPN network will generate five feature maps with different scales.
Without this model expansion, the classification operation can be described as the function:
	\begin{small}
		\begin{equation}
		\mathcal F_{i} = CONV_{FPN_i}(\mathcal X)
		\label{eq: FCOS feature extraction}
		\end{equation}
	\end{small}
	\begin{small}
		\begin{equation}
		\mathcal Y = CONV_{Cls\_Old}(CONV_{CH}(\ \mathcal F_i)\ )
		\label{eq: FCOS classification}
		\end{equation}
	\end{small}

In our method, for each feature map ($\mathcal F_{i}$) from FPN, we adapt it individually by a dilatable module $CONV_{DM\_FPN_i}$.
For all five feature maps from `Cls-Head CONV Tower', similar to FCOS, we adapt them by a shared dilatable module $CONV_{DM\_CH}$.
Thus, with model expansion, the classification operation can be described by the function:
	\begin{small}
		\begin{equation}
		\mathcal F_{i}^{'} = CONV_{DM\_FPN_i}(\mathcal F_{i} )
		\label{eq: Dilatable feature extraction}
		\end{equation}
	\end{small}
	\begin{small}
		\begin{equation}
		\mathcal Y^{'} = CONV_ {Cls\_New}(\ CONV_{DM\_CH}(\ CONV_{CH}(\mathcal F_{i}^{'})\ )\ )
		\label{eq: Dilatable prediction}
		\end{equation}
	\end{small}

In addition, for object detection, multiple objects can exist in one input image, and under incremental scenarios, these objects can come from both previous and current tasks. 
Thus, there is no task identity for the input image. 
Although we create disjoint classification heads for new classes, the new classification head will work together with all previous classification heads to detect old class objects as well as new class objects. 
The disjoint classification heads for old and new classes can be regarded as a cascaded unified classification head.
	
\subsection{DIODE}
\label{sec: DIODE}
To sum up, during each new task learning phase, the EWC regularization is applied with constraints on the parameters for ResNet, FPN, and `Cls-Head CONV Tower'.
At the same time, the dilatable module is added on each FPN feature map and the output from `Cls-Head CONV Tower'.
The parameters on newly created dilatable modules and new class classification head are updated only by the original training loss of the FCOS detector.
Parameters on the regression heads (`Reg-Head CONV Tower', `Center-ness', and `Regression') are also trained only by the standard FCOS training loss.
As mentioned in Section \ref{sec: Constrained EWC Module}, the regression outputs are partly trained only at positive pixels where annotations exist, so standard normal training will not affect the old object regression.
The parameters on old class classification heads are fixed since they are only related to the old tasks.
In addition, the neural network is normally overloaded with parameters from initial task learning. 
To adequately utilize the existing parameters, in all our experimental settings, we do not expand the model until the second incremental step.
If there is a one-step incremental condition that requires a large model capacity, the model can be simply expanded from the first incremental step.
		
\section{Experiments}
\label{sec: Experiments}
In this section, we perform experiments on different multi-step incremental scenarios on both the VOC and COCO datasets.
An ablation study is performed to validate the elements of our method.
	
\subsection{Dataset and Evaluation Metric}
We use the benchmark datasets PASCAL VOC 2007 \cite{everingham2010pascal} and COCO 2014 \cite{lin2014microsoft} for our experiments.
For experiments on VOC, the train and validation sets are used for training, and the test set is used for testing.
For experiments on COCO, the train and valminusminival sets are used for training, and the minival set is used for testing.
During incremental learning, the training data are provided by the following rules:
1) The new categories are continually added in alphabetical order.
2) Under multi-step incremental learning, all classes trained before are regarded as old classes.
3) Only the training data for the new classes are available.
4) The annotations for old class objects on new class data are not provided.
	
For the evaluation metric, mean average precision (mAP) at 0.5 Intersection over Union (IoU) is used for both datasets.
As per the common research protocol, mAP weighted across various IoUs from 0.5 to 0.95 is used as an additional metric for COCO.

\begin{table*}[tpb]
	\centering
	\scriptsize{
		\caption{Experimental results for \emph{the three-step incremental adding five new classes at a time} protocol on the VOC dataset.
			(a-b) is the first step normal training for categories a to b and +(c-d) is the incremental learning for categories c to d.}
		\begin{tabular}{|l|c|c|c|c|c|}
			\hline
			& (1-5) & +(6-10) & +(11-15) & +(16-20) & (1-20) \\
			\hline
			Directly fine-tuning (Catastrophic forgetting) & \multirow{11}*{70.6\%} & 37.2\% & 23.0\% & 12.4\% &  \multirow{11}*{71.6\%} \\
			\cline{1-1} \cline{3-5}
			LwF \cite{li2017learning} &  & 34.3\% & - & - & \\
			\cline{1-1} \cline{3-5}
			EWC \cite{kirkpatrick2017overcoming} &  & 37.5\% & 24.3\% & 13.0\% & \\
			\cline{1-1} \cline{3-5}
			Online EWC \cite{schwarz2018progress} &  & 37.5\% & 24.3\% & 13.9\% & \\
			\cline{1-1} \cline{3-5}
			MAS \cite{aljundi2018memory} &  & 40.5\% & 27.8\% & 13.1\% & \\
			\cline{1-1} \cline{3-5}
			Directly fine-tuning w pseudo annotation &  & 58.9\% & 52.5\% & 39.1\% & \\
			\cline{1-1} \cline{3-5}
			EWC \cite{kirkpatrick2017overcoming} w pseudo annotation &  & 60.3\% & 51.9\% & 39.2\% & \\
			\cline{1-1} \cline{3-5}
			Online EWC \cite{schwarz2018progress} w pseudo annotation &  & 60.3\% & 51.3\% & 38.3\% & \\
			\cline{1-1} \cline{3-5}
			MAS \cite{aljundi2018memory} w pseudo annotation &  & 61.3\% & 52.8\% & 38.2\% & \\
			\cline{1-1} \cline{3-5}
			IncDet \cite{liu2020incdet} &  & 61.1\% & 54.4\% & 46.1\% & \\
			\cline{1-1} \cline{3-5}
			DIODE (Ours) &  & \textbf{64.1\%} & \textbf{56.5\%} & \textbf{47.7\%} &  \\
			\hline
		\end{tabular}	
		\label{tab: 5_5_5_5_FCOS}
	}
	\\~
	\centering
	\scriptsize{
		\caption{Experimental results for \emph{the four-step incremental learning four new classes at a time} protocol on the VOC dataset.}
		\begin{tabular}{|l|c|c|c|c|c|c|c|}
			\hline
			& (1-4) & +(5-8) & +(9-12) & +(13-16) & +(17-20) & (1-20) \\
			\hline
			Directly fine-tuning (Catastrophic forgetting) & \multirow{4}*{71.9\%} & 37.2\% & 19.7\% & 15.8\% & 9.3\% & \multirow{4}*{71.6\%} \\
			\cline{1-1} \cline{3-6}
			Online EWC \cite{schwarz2018progress} & & 37.5\% & 19.3\% & 16.9\% & 9.9\% & \\
			\cline{1-1} \cline{3-6}
			IncDet \cite{liu2020incdet} & & \textbf{57.9\%} & 49.0\% & 42.5\% & 37.2\% & \\
			\cline{1-1} \cline{3-6}
			DIODE (Ours) & & 57.3\% & \textbf{49.3\%} & \textbf{47.3\%} & \textbf{42.2\%} & \\
			\hline
		\end{tabular}	
		\label{tab: 4_4_4_4_4_FCOS}
	}
	\\~
	\centering
	\scriptsize{
		\caption{Experimental results for \emph{the five-step incremental learning two new classes at a time} protocol on the VOC dataset.}
		\begin{tabular}{|l|c|c|c|c|c|c|c|}
			\hline
			& (1-10) & +(11-12) & +(13-14) & +(15-16) & +(17-18) & +(19-20) & (1-20) \\
			\hline
			Directly fine-tuning (Catastrophic forgetting) & \multirow{4}*{73.4\%} & 14.9\% & 7.8\% & 7.8\% & 4.1\% & 6.1\% & \multirow{4}*{71.6\%} \\
			\cline{1-1} \cline{3-7}
			Online EWC \cite{schwarz2018progress} & & 29.8\% & 14.7\% & 8.3\% & 4.7\% & 6.5\% & \\
			\cline{1-1} \cline{3-7}
			IncDet \cite{liu2020incdet} & & 64.2\% & 54.0\% & 44.6\% & 39.1\% & 38.4\% & \\
			\cline{1-1} \cline{3-7}
			DIODE (Ours) & & \textbf{64.4\%} & \textbf{54.8\%} & \textbf{50.6\%} & \textbf{44.1\%} & \textbf{44.1\%} & \\
			\hline
		\end{tabular}	
		\label{tab: 10_2_2_2_2_2_FCOS}
	}
	\\~
	\centering
	\scriptsize{
		\caption{Experimental results for \emph{the five-step incremental learning one new class at a time} protocol on the VOC dataset.}
		\begin{tabular}{|l|c|c|c|c|c|c|c|}
			\hline
			& (1-15) & +(16) & +(17) & +(18) & +(19) & +(20) & (1-20)\\
			\hline
			Directly fine-tuning (Catastrophic forgetting) & \multirow{4}*{73.7\%} & 21.1\% & 10.3\% & 4.8\% & 1.4\% & 2.5\% & \multirow{4}*{71.6\%} \\
			\cline{1-1} \cline{3-7}
			Online EWC \cite{schwarz2018progress} & & 23.3\% & 10.6\% & 5.9\% & 4.0\% & 2.9\% & \\
			\cline{1-1} \cline{3-7}
			IncDet \cite{liu2020incdet} & & 64.4\% & 53.7\% & 50.0\% & 44.4\% & 41.0\% & \\
			\cline{1-1} \cline{3-7}
			DIODE (Ours) & & \textbf{67.2\%} & \textbf{57.0\%} & \textbf{52.5\%} & \textbf{46.2\%} & \textbf{45.6\%} & \\
			\hline
		\end{tabular}	
		\label{tab: 15_1_1_1_1_1_FCOS}
	}
\end{table*}
	
\subsection{Model Implementation Details}
We build our method based on FCOS \cite{tian2019fcos} using the official implementation with PyTorch.
ResNet-50 \cite{he2016deep} and FPN \cite{lin2017focal} are used as the feature extraction backbone.
To make fair comparisons with other methods, we reproduce EWC \cite{kirkpatrick2017overcoming}, online EWC \cite{schwarz2018progress}, LwF \cite{li2017learning}, MAS \cite{aljundi2018memory}, and IncDet \cite{liu2020incdet} methods on FCOS following the methodology mentioned in their papers.
All the models are trained using the same training strategy.
For the first training step, the network is trained by 40k iterations for VOC and 400k iterations for COCO.
In the following incremental steps, the network is trained for 10k iterations when only one class is added and the same number of iterations as the first step if multiple classes are added in a single step.
The learning rate is set to 0.001, decaying to 0.0001 after 30k iterations.

\begin{table*}[tpb]
	\centering
	\scriptsize{
		\caption{Experimental results for \emph{the four-step incremental learning ten new classes at a time} protocol on the COCO dataset.}
		\begin{tabular}{|l|c|c|c|c|c|c|c|}
			\hline
			& (1-40) & +(41-50) & +(51-60) & +(61-70) & +(71-80) & (1-80)\\
			\hline
			IncDet \cite{liu2020incdet} (mAP@.5) & \multirow{2}*{51.3\%} & 42.5\% & 36.5\% & 32.8\% & 32.0\% & \multirow{2}*{47.0\%} \\
			\cline{1-1} \cline{3-6}
			DIODE (Ours) (mAP@.5) & & \textbf{43.0\%} & \textbf{39.0\%} & \textbf{37.5\%} & \textbf{37.4\%} & \\
			\hline
			\hline
			IncDet \cite{liu2020incdet} (mAP@[.5, .95]) & \multirow{2}*{32.1\%} & 26.1\% & 22.9\% & 20.5\% & 19.8\% & \multirow{2}*{29.3\%} \\
			\cline{1-1} \cline{3-6}
			DIODE (Ours) (mAP@[.5, .95]) & & \textbf{26.5\%} & \textbf{24.0\%} & \textbf{22.9\%} & \textbf{22.9\%} & \\
			\hline
		\end{tabular}	
		\label{tab: 40_10_10_10_10_FCOS}
	}
	\\~
	\centering
	\scriptsize{
		\caption{Experimental results for \emph{the four-step incremental learning five new classes at a time} protocol on the COCO dataset.}
		\begin{tabular}{|l|c|c|c|c|c|c|}
			\hline
			& (1-60) & +(61-65) & +(66-70) & +(71-75) & +(76-80) & (1-80) \\
			\hline
			IncDet \cite{liu2020incdet} (mAP@.5) & \multirow{2}*{47.7\%} & \textbf{44.9\%} & 41.5\% & 40.3\% & 39.4\% & \multirow{2}*{47.0\%} \\
			\cline{1-1} \cline{3-6}
			DIODE (Ours) (mAP@.5) & & 44.5\% & \textbf{42.0\%} & \textbf{41.0\%} & \textbf{40.9\%} & \\
			\hline
			\hline
			IncDet \cite{liu2020incdet} (mAP@[.5, .95]) & \multirow{2}*{29.9\%} & \textbf{28.1\%} & \textbf{25.8\%} & 24.8\% & 24.4\% & \multirow{2}*{29.3\%} \\
			\cline{1-1} \cline{3-6}
			DIODE (Ours) (mAP@[.5, .95]) & & 27.5\% & 25.4\% & \textbf{25.0\%} & \textbf{24.9\%} & \\
			\hline
		\end{tabular}	
		\label{tab: 60_5_5_5_5_FCOS}
	}
	\\~
	\centering
	\scriptsize{
		\caption{Ablation study for \emph{the three-step incremental leaning five new classes at a time} protocol on the VOC dataset.}
		\begin{tabular}{|l|c|c|c|c|c|}
			\hline
			& (1-5) & +(6-10) & +(11-15) & +(16-20) & (1-20) \\
			\hline
			Directly fine-tuning w pseudo annotation & \multirow{5}*{70.6\%} & 58.9\% & 52.5\% & 39.1\% & \multirow{5}*{71.6\%} \\
			\cline{1-1} \cline{3-5}
			Online EWC \cite{schwarz2018progress} w pseudo annotation &  & 60.3\% & 51.3\% & 38.3\% & \\
			\cline{1-1} \cline{3-5}
			Constrained EWC w pseudo annotation &  & \textbf{64.1\%} & 54.7\% & 42.0\% & \\
			\cline{1-1} \cline{3-5}
			IncDet \cite{liu2020incdet} (including pseudo annotation) w model expansion  &  & 61.1\% & 54.0\% & 46.6\% & \\
			\cline{1-1} \cline{3-5}
			Constrained EWC w pseudo annotation w model expansion (DIODE) &  & \textbf{64.1\%} & \textbf{56.5\%} & \textbf{47.7\%} & \\
			\hline
		\end{tabular}	
		\label{tab: ablation_study}
	}
	\\~
	\centering
	\scriptsize{
		\caption{Analysis for the capacity of the model parameters.
			Using FCOS (ResNet50 + FPN) as the backbone, the percentage shows the ratio of the increased parameters for our DIODE method towards the original model.}
		\begin{tabular}{|l|c|c|c|c|c|c|c|}
			\hline
			& Task 0 & Task 1 & Task 2 & Task 3 & Task 4 & Task 5 \\
			\hline
			Task description & \tabincell{c}{normal \\ training} & \tabincell{c}{incremental \\ step 1} & \tabincell{c}{incremental \\ step 2} & \tabincell{c}{incremental \\ step 3} & \tabincell{c}{incremental \\ step 4} & \tabincell{c}{incremental \\ step 5} \\
			\hline
			Increased amount of parameters & 0.0\% & 0.0\% & 1.2\% & 2.5\% & 3.7\% & 4.9\% \\
			\hline
		\end{tabular}	
		\label{tab: parameter_analysis}
	}
\end{table*}
	
\subsection{Experiments on the VOC Dataset}
We perform four different multi-step incremental learning tasks on the VOC dataset.
Table \ref{tab: 5_5_5_5_FCOS} shows the experimental results for three-step incremental adding five new classes at a time scenario.
LwF fails from the second incremental step because it directly uses the outputs of the source model to guide the target model.
As the regression output of FCOS is partially trained, it contains strong noise which will not only fail to preserve old class information but also hinder new class learning.
With the help of pseudo annotations, the performance of EWC, online EWC, and MAS has been largely improved which validates the effectiveness of the pseudo annotation mentioned in Section \ref{sec: Pseudo Annotation}.
Even with the help of pseudo annotations, our method still outperforms EWC, online EWC, and MAS on all three incremental steps by an average of 5.7\%, 6.2\%, and 5.4\% mAP, respectively.
Our method also achieves better results than the current state-of-the-art method, IncDet, on all three steps by an average of 2.3\% mAP.

Table \ref{tab: 4_4_4_4_4_FCOS} shows the results for four-step incremental adding four new classes at a time scenario.
For all four incremental steps, our method outperforms online EWC by an average of 28.1\% mAP.
Although for the first two incremental steps, our method achieves similar results as IncDet, by the third and fourth incremental steps, our method significantly outperforms IncDet by 4.8\% and 5.0\% mAP, respectively.
This proves that with the aid of the dilatable module, our method is more robust and retains plasticity under multi-step learning.

Table \ref{tab: 10_2_2_2_2_2_FCOS} and Table \ref{tab: 15_1_1_1_1_1_FCOS} show the results for five-step incremental adding two and one new classes at a time scenarios.
With both settings, our method shows better results than online EWC and IncDet on all five incremental learning steps.
Under adding two-class at a time protocol, our method outperforms online EWC and IncDet by an average of 38.8\% and 3.5\% mAP, respectively.
Comparing with IncDet, our improvements become more obvious with the continually adding of new tasks in longer sequences.
Under adding one-class at a time protocol, our method largely outperforms online EWC and IncDet by an average of 44.4\% and 3.0\% mAP, respectively.
We also find that for different incremental settings, the improvements of our method vary.
We conjecture this is because the levels of learning difficulty are various for different incremental settings.
For incremental object detection, the learning difficulty highly depends on the amount of training data, the percentage of missing annotation for old classes, the ratio between the old and new classes, and the similarity between the old and new classes.
Note that the pseudo annotation used in our model has largely alleviated the missing annotation problem.
	
\subsection{Experiments on the COCO Dataset}
We perform two different multi-step incremental learning tasks on the COCO dataset.
Table \ref{tab: 40_10_10_10_10_FCOS}  and Table \ref{tab: 60_5_5_5_5_FCOS} show the results for four-step incremental adding ten and five new classes at a time scenarios, respectively.
Under adding ten-class at a time protocol, our method outperforms IncDet at all four steps by an average of 3.1\% mAP at 0.5 IoU and 1.8\% mAP at 0.5 to 0.95 IoU.
Similar to the results from the experiments on the VOC dataset, with the continual adding of new tasks, improvements become more obvious.
Under adding five-class at a time protocol, for the fourth step, our method outperforms IncDet by 1.5\% mAP at 0.5 IoU and 0.5\% mAP at 0.5 to 0.95 IoU.

To sum up, under six multi-step incremental scenarios (four on VOC and two on COCO), we outperform the state-of-the-art in almost all experiments.
The improvement is especially obvious on the last step, 1.6\%, 5\%, 5.7\% and 4.6\%, respectively compared to IncDet on four different VOC settings and 5.4\% and 1.5\% improvement compared to IncDet on two different COCO settings.
Due to the limitations of GPU, we lack the ability to do very large-scale multi-step experiments. 
But the improvement on the last step of six different settings demonstrates the effectiveness of our method on multi-step scenarios.
	
\subsection{Ablation Study}
To validate the effectiveness of each part of our proposed method, we perform an ablation study under adding five new classes at a time protocol on the VOC dataset.
There are three incremental learning steps in total.
Table \ref{tab: ablation_study} shows the experimental results.	
Our method mainly consists of two parts: constrained EWC module and dilatable module.
The constrained EWC module helps to preserve the old task performance.
Observing from Table \ref{tab: ablation_study}, directly applying the online EWC method to FCOS fails to preserve old model information.
The performance of online EWC with pseudo annotation is even worse than directly fine-tuning with pseudo annotation on the last two steps.
As mentioned in Section \ref{sec: Constrained EWC Module}, there are huge differences among the importance for different parameters and the extremely large importance are usually parameters from the bounding box regression head or classification head.
Directly applying EWC to the FCOS detector will make the regularization loss focus on detection heads and ignore the backbone which contains most of the old task knowledge.
One solution is to constrainedly apply EWC regularization only on the backbone parameters of the FCOS detector.
With the help of constrained EWC module, our method outperforms directly fine-tuning and online EWC on all three steps with an average of 3.4\% and 3.6\% mAP, respectively.
	
Then to compensate the negative side of the EWC module (inhibit new task learning) for long-range incremental learning, the dilatable module is applied.
As mentioned in Section \ref{sec: DIODE}, to make full use of the existing parameters, the dilatable module is not applied until the second incremental step.
With the help of the dilatable module, the performance of the model is further improved by 1.8\% and 5.7\% mAP at the second and third incremental step, respectively.
We also try to combine the dilatable module with the IncDet EWC method instead of the constrained EWC method.
However, observing from Table \ref{tab: ablation_study} and Table \ref{tab: 5_5_5_5_FCOS}, this combination does not provide a better result than constrained EWC with the dilatable module or even individual IncDet.
We conjecture this is because the IncDet method uses Huber regularization to clip the gradients of the large values for certain parameters on the EWC regularization term.
Although this threshold helps to avoid the gradient explosion problem, it causes another problem - the original distribution of the parameter importance is modified which adversely affects the old knowledge preserving capability of EWC. 
This makes IncDet incompatible with model expansion since some of the important parameters for old classes do not acquire adequate attention and are still updated.
	
Table \ref{tab: parameter_analysis} shows the percentage of increased parameters for our DIODE method.
To make full use of the existing neurons on the model, in our method, we do not expand the network structure until the second step of incremental learning.
After the first step of incremental learning, for each new task, newly added convolutional layers with kernel size one are applied on FPN and `Cls-Head CONV Tower' features to adapt them to the new task learning.
We try to expand the model by convolutional layers with different kernel sizes and find that using kernel size one provides the best performance with the least amount of parameters added.
As only a limited number of task-specific parameters are introduced, the capacity of the model increases very slowly.
After incremental learning five tasks, the parameters of the model are only increased by 4.9\%.
	
\subsection{Hyper-parameter Setting}
\begin{table}[tpb]
	\centering
	\scriptsize{
		\caption{Hyper-parameter ($\lambda$) values used for \emph{the three-step incremental learning five new classes at a time} protocol on the VOC dataset.}
		\begin{tabular}{|l|c|c|c|}
			\hline
			& +(6-10) & +(11-15) & +(16-20) \\
			\hline
			LwF \cite{li2017learning} & $1$ & - & - \\
			\hline
			EWC \cite{kirkpatrick2017overcoming} w/wo pseudo & $10^4$ & $10^3$ & $10^3$ \\
			\hline
			Online EWC \cite{schwarz2018progress} w/wo pseudo & $10^4$ & $10^3$ & $10^3$ \\
			\hline
			MAS \cite{schwarz2018progress} w/wo pseudo & $10$ & $10$ & $10$ \\
			\hline
			IncDet \cite{liu2020incdet} & $5 \times 10^3$ & $5 \times 10^3$ & $5 \times 10^3$ \\
			\hline
			DIODE (Ours) & $10^6$ & $10^4$ & $10^4$ \\
			\hline
		\end{tabular}	
		\label{tab: 5_5_5_5_FCOS_hyperparameter}}
\end{table}

As mentioned in \cite{aljundi2018memory, hsu2018re}, the performance of EWC type methods is prone to the choice of the regularization coefficient ($\lambda$), since $\lambda$ is a trade-off between the allowed forgetting and the new task loss.
We find that normally the best $\lambda$ is acquired just before gradient explosion happens where the regularization loss is adequately applied to consolidate the important old class parameters. 
Therefore, for all incremental steps, we use the critical point before gradient explosion as the $\lambda$ value.

Table \ref{tab: 5_5_5_5_FCOS_hyperparameter} shows the $\lambda$ values for adding five new classes at a time protocol on the VOC dataset.
For IncDet, as it employs Huber regularization to clip the gradient, the choice of $\lambda$ will not cause gradient explosion.
Referencing the hyper-parameter values mentioned in their paper, we use grid search on the first incremental step to find the best $\lambda$ for IncDet.
For the original EWC, the high importance values on the heads of the detector can easily cause gradient explosion and restricts the choice of high $\lambda$ value.
The use of small $\lambda$ values makes the EWC method unable to effectively slow the update of the parameters that are important for the old tasks.
Thus, it cannot properly preserve old task knowledge.
Similar to the EWC method, the MAS method also suffers from gradient explosion and is restricted to small $\lambda$ values which impede old knowledge preservation.
For our DIODE method, we adaptively apply EWC regularization only on the backbone parameters of the FCOS detector.
Thus, our method removes those extra-large importance parameters of the detector heads which also contain minor old task knowledge.
Due to this reason, in our method, a much larger $\lambda$ value can be used to better preserve old model information.
	
\section{Conclusion}
\label{sec: Conclusion}
In this paper, we focus on designing a dilatable incremental object detector (DIODE) for multi-step incremental detection.
We first analyze the incremental classification method EWC with respect to the anchor-free object detector FCOS.
Then according to the characteristics of the importance matrix generated by EWC on the FCOS structure, we adaptively apply the EWC method to the backbone network of the FCOS detector to preserve old task knowledge.
After that, during multi-step incremental learning, to make full use of the existing neurons and promote the learning of new tasks, from the second step of incremental learning, we expand the model structure with a modest number of parameters.
Experiments on the benchmark datasets demonstrate the effectiveness of our method.

\section*{Acknowledgments}
This research was funded by the Australian Government through the Australian Research Council and Sullivan Nicolaides Pathology under Linkage Project LP160101797.

{\small
	\bibliographystyle{IEEEtranS}
	\bibliography{egbib}

\begin{thebibliography}{10}
\providecommand{\url}[1]{#1}
\csname url@samestyle\endcsname
\providecommand{\newblock}{\relax}
\providecommand{\bibinfo}[2]{#2}
\providecommand{\BIBentrySTDinterwordspacing}{\spaceskip=0pt\relax}
\providecommand{\BIBentryALTinterwordstretchfactor}{4}
\providecommand{\BIBentryALTinterwordspacing}{\spaceskip=\fontdimen2\font plus
\BIBentryALTinterwordstretchfactor\fontdimen3\font minus
  \fontdimen4\font\relax}
\providecommand{\BIBforeignlanguage}[2]{{%
\expandafter\ifx\csname l@#1\endcsname\relax
\typeout{** WARNING: IEEEtranS.bst: No hyphenation pattern has been}%
\typeout{** loaded for the language `#1'. Using the pattern for}%
\typeout{** the default language instead.}%
\else
\language=\csname l@#1\endcsname
\fi
#2}}
\providecommand{\BIBdecl}{\relax}
\BIBdecl

\bibitem{abati2020conditional}
D.~Abati, J.~Tomczak, T.~Blankevoort, S.~Calderara, R.~Cucchiara, and B.~E.
  Bejnordi, ``Conditional channel gated networks for task-aware continual
  learning,'' in \emph{Proceedings of the IEEE/CVF Conference on Computer
  Vision and Pattern Recognition}, 2020, pp. 3931--3940.

\bibitem{aljundi2018memory}
R.~Aljundi, F.~Babiloni, M.~Elhoseiny, M.~Rohrbach, and T.~Tuytelaars, ``Memory
  aware synapses: Learning what (not) to forget,'' in \emph{Proceedings of the
  European Conference on Computer Vision (ECCV)}, 2018, pp. 139--154.

\bibitem{amari1998natural}
S.-I. Amari, ``Natural gradient works efficiently in learning,'' \emph{Neural
  computation}, vol.~10, no.~2, pp. 251--276, 1998.

\bibitem{castro2018end}
F.~M. Castro, M.~J. Mar{\'\i}n-Jim{\'e}nez, N.~Guil, C.~Schmid, and K.~Alahari,
  ``End-to-end incremental learning,'' in \emph{Proceedings of the European
  Conference on Computer Vision (ECCV)}, 2018, pp. 233--248.

\bibitem{8851980}
L.~{Chen}, C.~{Yu}, and L.~{Chen}, ``A new knowledge distillation for
  incremental object detection,'' in \emph{2019 International Joint Conference
  on Neural Networks (IJCNN)}, 2019, pp. 1--7.

\bibitem{everingham2010pascal}
M.~Everingham, L.~Van~Gool, C.~K. Williams, J.~Winn, and A.~Zisserman, ``The
  pascal visual object classes (voc) challenge,'' \emph{International journal
  of computer vision}, vol.~88, no.~2, pp. 303--338, 2010.

\bibitem{fernando2017pathnet}
C.~Fernando, D.~Banarse, C.~Blundell, Y.~Zwols, D.~Ha, A.~A. Rusu, A.~Pritzel,
  and D.~Wierstra, ``Pathnet: Evolution channels gradient descent in super
  neural networks,'' \emph{arXiv preprint arXiv:1701.08734}, 2017.

\bibitem{Girshick2015}
R.~Girshick, ``Fast r-cnn,'' in \emph{Proceedings of the IEEE international
  conference on computer vision}, 2015, pp. 1440--1448.

\bibitem{goodfellow2013empirical}
I.~J. Goodfellow, M.~Mirza, D.~Xiao, A.~Courville, and Y.~Bengio, ``An
  empirical investigation of catastrophic forgetting in gradient-based neural
  networks,'' in \emph{Proceedings of International Conference on Learning
  Representations}, 2014.

\bibitem{Hao2019AnEA}
Y.~Hao, Y.~Fu, Y.-G. Jiang, and Q.~Tian, ``An end-to-end architecture for
  class-incremental object detection with knowledge distillation,'' \emph{2019
  IEEE International Conference on Multimedia and Expo (ICME)}, pp. 1--6, 2019.

\bibitem{hayashi2015labelling}
A.~Hayashi-Takagi, S.~Yagishita, M.~Nakamura, F.~Shirai, Y.~I. Wu, A.~L.
  Loshbaugh, B.~Kuhlman, K.~M. Hahn, and H.~Kasai, ``Labelling and optical
  erasure of synaptic memory traces in the motor cortex,'' \emph{Nature}, vol.
  525, no. 7569, pp. 333--338, 2015.

\bibitem{he2016deep}
K.~He, X.~Zhang, S.~Ren, and J.~Sun, ``Deep residual learning for image
  recognition,'' in \emph{Proceedings of the IEEE conference on computer vision
  and pattern recognition}, 2016, pp. 770--778.

\bibitem{hinton2015distilling}
G.~Hinton, O.~Vinyals, and J.~Dean, ``Distilling the knowledge in a neural
  network,'' \emph{NIPS Deep Learning and Representation Learning Workshop},
  2015.

\bibitem{hou2019learning}
S.~Hou, X.~Pan, C.~C. Loy, Z.~Wang, and D.~Lin, ``Learning a unified classifier
  incrementally via rebalancing,'' in \emph{Proceedings of the IEEE Conference
  on Computer Vision and Pattern Recognition}, 2019, pp. 831--839.

\bibitem{hsu2018re}
Y.-C. Hsu, Y.-C. Liu, A.~Ramasamy, and Z.~Kira, ``Re-evaluating continual
  learning scenarios: A categorization and case for strong baselines,''
  \emph{arXiv preprint arXiv:1810.12488}, 2018.

\bibitem{jung2020continual}
S.~Jung, H.~Ahn, S.~Cha, and T.~Moon, ``Continual learning with node-importance
  based adaptive group sparse regularization,'' \emph{Advances in Neural
  Information Processing Systems}, vol.~33, 2020.

\bibitem{kirkpatrick2017overcoming}
J.~Kirkpatrick, R.~Pascanu, N.~Rabinowitz, J.~Veness, G.~Desjardins, A.~A.
  Rusu, K.~Milan, J.~Quan, T.~Ramalho, A.~Grabska-Barwinska \emph{et~al.},
  ``Overcoming catastrophic forgetting in neural networks,'' \emph{Proceedings
  of the national academy of sciences}, vol. 114, no.~13, pp. 3521--3526, 2017.

\bibitem{li2019rilod}
D.~Li, S.~Tasci, S.~Ghosh, J.~Zhu, J.~Zhang, and L.~Heck, ``Rilod: near
  real-time incremental learning for object detection at the edge,'' in
  \emph{Proceedings of the 4th ACM/IEEE Symposium on Edge Computing}, 2019, pp.
  113--126.

\bibitem{li2017learning}
Z.~Li and D.~Hoiem, ``Learning without forgetting,'' \emph{IEEE transactions on
  pattern analysis and machine intelligence}, vol.~40, no.~12, pp. 2935--2947,
  2017.

\bibitem{lin2017focal}
T.-Y. Lin, P.~Goyal, R.~Girshick, K.~He, and P.~Doll{\'a}r, ``Focal loss for
  dense object detection,'' in \emph{Proceedings of the IEEE international
  conference on computer vision}, 2017, pp. 2980--2988.

\bibitem{lin2014microsoft}
T.-Y. Lin, M.~Maire, S.~Belongie, J.~Hays, P.~Perona, D.~Ramanan,
  P.~Doll{\'a}r, and C.~L. Zitnick, ``Microsoft coco: Common objects in
  context,'' in \emph{European conference on computer vision}.\hskip 1em plus
  0.5em minus 0.4em\relax Springer, 2014, pp. 740--755.

\bibitem{liu2020incdet}
L.~Liu, Z.~Kuang, Y.~Chen, J.-H. Xue, W.~Yang, and W.~Zhang, ``Incdet: in
  defense of elastic weight consolidation for incremental object detection,''
  \emph{IEEE transactions on neural networks and learning systems}, 2020.

\bibitem{mallya2018packnet}
A.~Mallya and S.~Lazebnik, ``Packnet: Adding multiple tasks to a single network
  by iterative pruning,'' in \emph{Proceedings of the IEEE Conference on
  Computer Vision and Pattern Recognition}, 2018, pp. 7765--7773.

\bibitem{mccloskey1989catastrophic}
M.~McCloskey and N.~J. Cohen, ``Catastrophic interference in connectionist
  networks: The sequential learning problem,'' in \emph{Psychology of Learning
  and Motivation}.\hskip 1em plus 0.5em minus 0.4em\relax Elsevier, 1989,
  vol.~24, pp. 109--165.

\bibitem{peng2020faster}
C.~Peng, K.~Zhao, and B.~C. Lovell, ``Faster ilod: Incremental learning for
  object detectors based on faster rcnn,'' \emph{Pattern Recognition Letters},
  vol. 140, pp. 109--115, 2020.

\bibitem{perez2020incremental}
J.-M. Perez-Rua, X.~Zhu, T.~M. Hospedales, and T.~Xiang, ``Incremental few-shot
  object detection,'' in \emph{Proceedings of the IEEE/CVF Conference on
  Computer Vision and Pattern Recognition}, 2020, pp. 13\,846--13\,855.

\bibitem{rebuffi2017icarl}
S.-A. Rebuffi, A.~Kolesnikov, G.~Sperl, and C.~H. Lampert, ``icarl: Incremental
  classifier and representation learning,'' in \emph{Proceedings of the IEEE
  Conference on Computer Vision and Pattern Recognition}, 2017, pp. 2001--2010.

\bibitem{ren2015faster}
S.~Ren, K.~He, R.~Girshick, and J.~Sun, ``Faster r-cnn: Towards real-time
  object detection with region proposal networks,'' in \emph{Advances in neural
  information processing systems}, 2015, pp. 91--99.

\bibitem{rosenfeld2018incremental}
A.~Rosenfeld and J.~K. Tsotsos, ``Incremental learning through deep
  adaptation,'' \emph{IEEE transactions on pattern analysis and machine
  intelligence}, 2018.

\bibitem{rusu2016progressive}
A.~A. Rusu, N.~C. Rabinowitz, G.~Desjardins, H.~Soyer, J.~Kirkpatrick,
  K.~Kavukcuoglu, R.~Pascanu, and R.~Hadsell, ``Progressive neural networks,''
  \emph{arXiv preprint arXiv:1606.04671}, 2016.

\bibitem{schwarz2018progress}
J.~Schwarz, W.~Czarnecki, J.~Luketina, A.~Grabska-Barwinska, Y.~W. Teh,
  R.~Pascanu, and R.~Hadsell, ``Progress \& compress: A scalable framework for
  continual learning,'' in \emph{International Conference on Machine
  Learning}.\hskip 1em plus 0.5em minus 0.4em\relax PMLR, 2018, pp. 4528--4537.

\bibitem{shmelkov2017incremental}
K.~Shmelkov, C.~Schmid, and K.~Alahari, ``Incremental learning of object
  detectors without catastrophic forgetting,'' in \emph{Proceedings of the IEEE
  International Conference on Computer Vision}, 2017, pp. 3400--3409.

\bibitem{singh2020calibrating}
P.~Singh, V.~K. Verma, P.~Mazumder, L.~Carin, and P.~Rai, ``Calibrating cnns
  for lifelong learning,'' \emph{Advances in Neural Information Processing
  Systems}, vol.~33, 2020.

\bibitem{tian2019fcos}
Z.~Tian, C.~Shen, H.~Chen, and T.~He, ``Fcos: Fully convolutional one-stage
  object detection,'' in \emph{Proceedings of the IEEE international conference
  on computer vision}, 2019, pp. 9627--9636.

\bibitem{yang2009stably}
G.~Yang, F.~Pan, and W.-B. Gan, ``Stably maintained dendritic spines are
  associated with lifelong memories,'' \emph{Nature}, vol. 462, no. 7275, pp.
  920--924, 2009.

\bibitem{wu2019large}
L.~W. Y. Y. Z. L. Y. G. Y.~F. Yue~Wu, Yinpeng~Chen, ``Large scale incremental
  learning,'' in \emph{Proceedings of the IEEE conference on computer vision
  and pattern recognition}, 2019.

\bibitem{zenke2017continual}
F.~Zenke, B.~Poole, and S.~Ganguli, ``Continual learning through synaptic
  intelligence,'' in \emph{Proceedings of the 34th International Conference on
  Machine Learning-Volume 70}.\hskip 1em plus 0.5em minus 0.4em\relax JMLR.
  org, 2017, pp. 3987--3995.

\bibitem{zhang2020class}
J.~Zhang, J.~Zhang, S.~Ghosh, D.~Li, S.~Tasci, L.~Heck, H.~Zhang, and C.-C.~J.
  Kuo, ``Class-incremental learning via deep model consolidation,'' in
  \emph{The IEEE Winter Conference on Applications of Computer Vision}, 2020,
  pp. 1131--1140.

\bibitem{zhao2020maintaining}
B.~Zhao, X.~Xiao, G.~Gan, B.~Zhang, and S.-T. Xia, ``Maintaining discrimination
  and fairness in class incremental learning,'' in \emph{Proceedings of the
  IEEE/CVF Conference on Computer Vision and Pattern Recognition}, 2020, pp.
  13\,208--13\,217.

\bibitem{zhou2019m2kd}
P.~Zhou, L.~Mai, J.~Zhang, N.~Xu, Z.~Wu, and L.~S. Davis, ``M2kd: Multi-model
  and multi-level knowledge distillation for incremental learning,''
  \emph{arXiv preprint arXiv:1904.01769}, 2019.

\bibitem{zhou2019objects}
X.~Zhou, D.~Wang, and P.~Kr{\"a}henb{\"u}hl, ``Objects as points,'' \emph{arXiv
  preprint arXiv:1904.07850}, 2019.

\end{thebibliography}
}	
\end{document}